\documentclass[sigconf]{acmart}
\def\BibTeX{{\rm B\kern-.05em{\sc i\kern-.025em b}\kern-.08emT\kern-.1667em\lower.7ex\hbox{E}\kern-.125emX}}

\setcopyright{none}
\renewcommand\footnotetextcopyrightpermission[1]{}
\copyrightyear{2020}
\acmYear{2020}
\setcopyright{acmlicensed}
\acmConference[MLG 2020]{16th International Workshop on Mining and Learning with Graphs}{Aug 24, 2020}{San Diego, CA, USA}
\acmBooktitle{MLG 2020: 16th International Workshop on Mining and Learning with Graphs, Aug 24, 2020, San Diego, CA, USA}
\acmPrice{15.00}
\acmDOI{10.1145/1122445.1122456}
\acmISBN{978-1-4503-9999-9/18/06}

\newcommand{\etal}{\emph{et al.}}
\newcommand{\eg}{\emph{e.g.}}
\newcommand{\ie}{\emph{i.e.}}
\newcommand{\ProjectName}[1]{{\small\textsc{UHGR}}}

\usepackage{amsmath}
\usepackage{amssymb}

\usepackage{graphicx}

\usepackage{changes}
\definechangesauthor[color=red]{is}
\definechangesauthor[color=blue]{js}
\definechangesauthor[color=green]{fd}

\usepackage[disable]{todonotes}

\begin{document}

%
\title{Unsupervised Hierarchical Graph Representation Learning by Mutual Information Maximization}

%

\author{Fei Ding}
\affiliation{%
  \institution{School of Computing, Clemson University}
  \city{Clemson}
  \country{USA}}
\email{feid@clemson.edu}

\author{Xiaohong Zhang}
\affiliation{%
  \institution{Department of Chemical and Biomolecular Engineering, Clemson University}
  \city{Clemson}
  \country{USA}
}
\email{xiaohoz@clemson.edu}

\author{Justin Sybrandt}
\affiliation{%
 \institution{School of Computing, Clemson University}
 \city{Clemson}
 \country{USA}}
 \email{jsybran@clemson.edu}
 
\author{Ilya Safro}
\affiliation{%
  \institution{School of Computing, Clemson University}
  \city{Clemson}
  \country{USA}}
\email{isafro@clemson.edu}

%
\renewcommand{\shortauthors}{Fei Ding, et al.}

%
\begin{abstract}
Graph representation learning based on graph neural networks (GNNs) can greatly improve the performance of downstream tasks, such as node and graph classification. However, the general GNN models do not aggregate node information in a hierarchical manner, and can miss key higher-order structural features of many graphs.
The hierarchical aggregation also enables the graph representations to be explainable. In addition, supervised graph representation learning requires labeled data, which is expensive and error-prone. To address these issues, we present an unsupervised graph representation learning method, Unsupervised Hierarchical Graph Representation (\ProjectName~), which can generate hierarchical representations of graphs. Our method focuses on maximizing mutual information between ``local'' and high-level ``global'' representations, which enables us to learn the node embeddings and graph embeddings without any labeled data. 
To demonstrate the effectiveness of the proposed method, we perform the node and graph classification using the learned node and graph embeddings. The results show that the proposed method achieves comparable results to state-of-the-art supervised methods on several benchmarks. In addition, our visualization of hierarchical representations indicates that our method can capture meaningful and interpretable clusters.\\
{\bf Reproducibility:} Our code and experimental data are available at this link\footnote{ \url{https://github.com/ifding/uhgr}}.
\end{abstract}

%
%


%

%
\maketitle

\section{Introduction}
Graph representation learning has been used in many domains that are related to graph-structured data, including bioinformatics~\cite{fout2017protein}, chemoinformatics~\cite{jin2017predicting,schutt2017schnet}, social networks~\cite{fang2016community} and cyber-security~\cite{vu2014cybersage}. There are two important tasks in graph analysis, \ie, label predictions on nodes and graphs. For instance, in the study of chemical molecules, researchers apply graph classification~\cite{lusci2013deep,lee2018graph,ying2018hierarchical} to help discover chemical properties of new molecule
\todo{really awkward sentence structure. How about "For instance, scientists use graph classification to help..."}
\todo[color=green]{done}
by predicting labels of the molecule, where a molecule can be represented as a graph with the atom represented as nodes and chemical bond represented as edges.

Graph neural networks (GNNs) are applied to graph-based data to improve prediction performance due to their ability to learn high-level features by propagating, transforming, and aggregating neighborhood information across edges~\cite{gilmer2017neural,hamilton2017inductive}. There are various neighborhood aggregation methods to capture the structures and attributes of graphs, including the average aggregation~\cite{kipf2016semi}, generalized aggregation~\cite{hamilton2017inductive} and attention-based aggregation~\cite{velivckovic2018deep}. 
However, these techniques sometimes miss key structural features for large, sparse, and noisy real-world graphs. In these cases, the most valuable information is often contain in several small subgraphs, which conventional aggregations methods often struggle to capture.

To solve this problem, Lee \etal~\cite{lee2018graph} present the Graph Attention Model (GAM), which focuses on small parts of graphs in order to predict the labels of the entire graphs. In order to improve embedding quality, the GAM model also integrates global information from various parts of the graph via different random sets of nodes. \todo{shouldn't be sets (not set)?} \todo[inline, color=green]{done} This suggests that local and global information are both important in graph representation learning. 
In the analysis of real-world graphs, it is necessary to gather information from individual nodes and edges as well as the subgraphs of graph that represent discriminative patterns. Recently, Ying \etal~\cite{ying2018hierarchical} proposed a graph pooling module, \textsc{DiffPool}, to generate hierarchical representations of graphs for the purpose of graph classification. This mechanism allows GNNs to encode the local and global structural information to obtain the final graph representation. 
\todo{I think this entire section should occur in a "background" section. It doesn't really introduce what you're trying to propose, or why your method is important.}
\todo[inline, color=green]{done}
Although the above methods perform well in the graph classification task, they are task-specific and focus on supervised learning.
These methods depend highly on vast quantities of labeled graph data, which is often costly and error-prone in the real world. To address this problem, Veli{\u c}kovi{\' c} \etal~\cite{velivckovic2018deep} applies mutual information maximization to learn node representations of graph-structured inputs without using labelled data, and demonstrates competitive performance to supervised learning on several node classification benchmarks.

Inspired by this work, we propose a novel unsupervised learning method, Unsupervised Hierarchical Graph Representation (\ProjectName~), to learn hierarchical graph representations based on mutual information maximization, which includes node embeddings and graph embeddings. We summarize the main contributions as follows:
\begin{itemize}
\item We propose an unsupervised hierarchical graph representation learning method to capture the local and global structural information of arbitrary sized graphs, which does not depend on any task-specific information (e.g., class labels). This method is generic enough to be used in various scenarios such as node embedding and graph embedding. 
\item We demonstrate that the graph representations from the proposed model can achieve comparable node and graph classification performance to supervised baseline methods on real-world data sets. 
\item The proposed method can learn meaningful and interpretable clusters across different levels of coarseness based on the structural information of graphs, as demonstrated through our visualizations.

\end{itemize}

The remainder of this paper is organized as follows:  section~\ref{section:method} illustrates our proposed method; the discussions of experimental results are provided in section~\ref{section:experiments}; section~\ref{section:related} reviews the related work; finally, we discuss the conclusions in section~\ref{section:conclusion}.
\todo{You can remove this section for space.}\todo[inline, color=green]{The maximum length of papers is 16 pages, we only have 14 pages.}

\section{Proposed Method}\label{section:method}

\begin{figure*}
\includegraphics[width=0.9\textwidth]{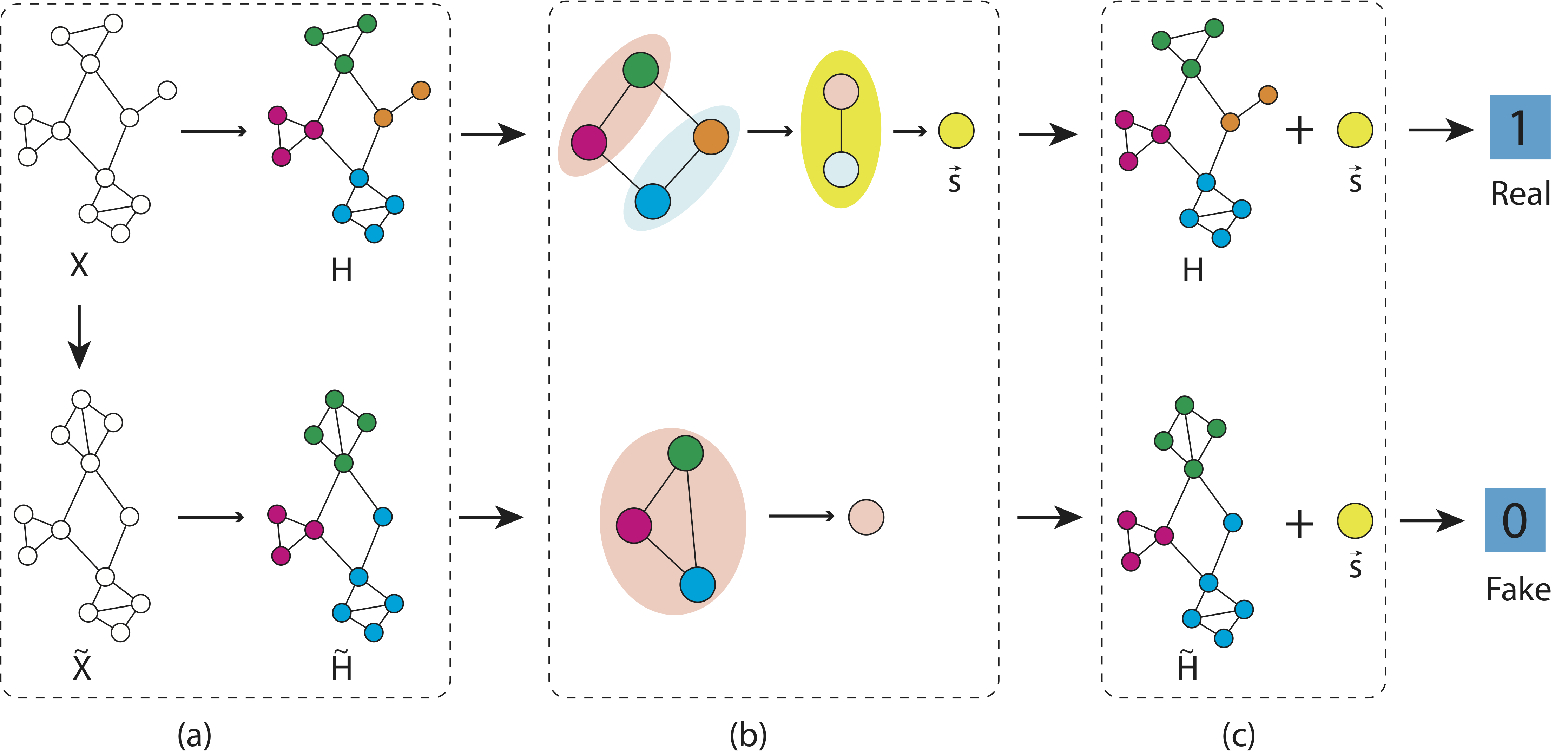}
\caption{The architecture of the Unsupervised Hierarchical Graph Representation (\ProjectName~) model. The left module (a) is an Encoder that creates the node representations H by exploiting the node feature X and the adjacency matrix A. The middle module (b) utilizes hierarchical graph pooling to create the graph summary $\Vec{s}$. The right module (c) is a Discriminator trained to discriminate if a pair of H and $\Vec{s}$ is generated from the same graph or not. }
\label{fig:overview}
\end{figure*}

Inspired by the recent success of unsupervised learning based \todo{what is based on what? first sentence should be more clear}\todo[inline, color=green]{unsupervised learning based on mutual information maximization} upon mutual information maximization~\cite{hjelm2018learning,velivckovic2018deep}, we propose a novel unsupervised embedding framework, \ProjectName~, to capture structural information and learn a hierarchical graph representation. This method is based on the maximization of mutual information between ``local'' features from neighbors of one node and high-level ``global'' features from the entire graph, which enables us to learn both node and graph representations.\todo{I find this to be a really unclear point. I do not know what sort of local or global features you're interested in? Roles, clusters, cliques, motifs?}\todo[inline, color=green]{Based on my understanding, the local or global features in machine learning are not like that. Roughly speaking, local features are from neighborhood of one node and global features are about entire graph.}
The proposed method utilizes the unsupervised learning method
to aggregate structural information to generate hierarchical representations. This unsupervised method makes the graph representations feasible for various downstream tasks, such as node and graph classification. Meanwhile, our method overcomes the shortcomings of previous studies that do not integrate different structural information of graphs well.
To evaluate our method, we apply the learned representations on the node and graph classification tasks, and compare the classification results with several baseline methods.

\subsection{Preliminaries}

The undirected graph $G = (X, A)$ is comprised of $n$ nodes, each with $f$ features.
Here, $X \in \mathbb{R}^{(n \times f)}$ where the original node features $\Vec{x_i}$ is read directly from files and represented by row $i$ of $X$. Furthermore, the adjacency matrix $A \in \{0, 1\}^{(n \times n)}$ contains a nonzero entry $A_{ij}$ to indicate an edge between nodes $i$ and  $j$.
The goal of this work is to create different levels of low-rank encodings of $G$, which we accomplish by training an encoder to cluster local parts of the graph and create more coarsened graphs, eventually output the final representation of the original graph\todo{enumerate the steps, hard to follow the sentence}. 
Each coarsened graph has its own node features and an adjacency matrix that are trainable. In order to train the encoder \todo{Don't capitalize Encoder, check everywhere}\todo[inline, color=green]{done} module, we apply a hierarchical approach where $G$ is repeatedly coarsened from $G_1=(H_1, A_1), \ldots, G_L(H_l, A_l)$ and $H$ represents the learned representations\todo{why do you replace notation X with H? $G=(X,A)$ with $G=(H,A)$}\todo[inline, color=green]{X:original node features and H: learned representations}. The $H_1$ and $A_1$ are from the original graph\todo{then why do you need $G_0$?}\todo[inline, color=green]{$G_0$ is the original graph as the input}, and the $H_l$ is the final graph representation of the original graph.
Following this scheme, the number of nodes in the successively coarsened graphs is non-increasing.
Because $H_i \in \mathbb{R}^{n_i \times f_i}$ represents the node embeddings of level\todo{Don't capitalize Level}\todo[inline, color=green]{done} \textit{i}, if $i < j$, then $n_i \leq n_j$. 
The feature vector corresponding to the coarse nodes is determined by a separate hierarchical level, $G_{i-1}\rightarrow \mathbb{R}^{f_i}$, which learns node embeddings of level $i$ from the previous level $i-1$ of coarseness.

This paper uses graph neural networks (GNNs) to create representations of the graphs at different levels, which is able to capture hierarchical structures and generate flexible graph embeddings. A key component of the proposed method is how to cluster partial parts of the graph and generate more coarsened graphs based on the output of GNNs without any labels. In the following parts, we outline the different modules of \ProjectName~ and illustrate how to learn hierarchical graph representations based on mutual information maximization.

\subsection{Encoder module}

The hierarchical encoder mainly depends on message-passing function $\mathcal{M}$. The message-passing function $\mathcal{M}$ is used to iteratively compute node representations from their neighborhood's features~\cite{ying2018hierarchical}:
\begin{equation}
  H^{(k)} = \mathcal{M} (A, H^{(k-1)}),
\end{equation}
where $H^{(k)}$ are the node embeddings of the k-th step from message-passing function $\mathcal{M}$, which depends on the adjacency matrix $A$ and the previous node embeddings $H^{(k-1)}$. At the initial step (k=1), $H^{(0)}$ is initialized by the original node features $X$. After K iterations, the module outputs the final node embeddings $Z = H^{(k)}$. The message-passing function $\mathcal{M}$ \todo{Do not start a sentence with math notation}\todo[color=green]{done} can be implemented by different types of GNNs. In this work, we consider two general GNNs: Graph Convolutional Networks (GCNs)~\cite{kipf2016semi} and Graph Attention Networks (GATs) ~\cite{velivckovic2017graph}.

\textbf{Graph Convolutional Networks}. GCNs implement $\mathcal{M}$ using the following rule:
\begin{equation}
  H^{(k)} = \sigma (\hat{D}^{-\frac{1}{2}} \hat{A} \hat{D}^{-\frac{1}{2}} H^{(k-1)} W^{(k-1)}),
\end{equation}
where $\hat{A} = A + I$ is the adjacency matrix with self-loops and $\hat{D}_{ii} = \sum_j \hat{A}_{ij}$ is the corresponding degree matrix. For the nonlinearity $\sigma$, we apply the parametric ReLU function~\cite{he2015delving}, and $W^{(k-1)}$ is a trainable weight matrix. 

\textbf{Graph Attention Networks}. GATs leverage self-attentional layers to set learnable weights to measure the importance of neighborhoods when aggregating feature information from node's neighbors. When computing new feature representation for a central node, each neighborhood receives a different weight by measuring the relation between its feature vector and the central node's vector. Node \textit{i} and its neighborhood node \textit{j} have the following relations: 
\begin{equation}
  e_{ij} = a (W\Vec{x_i}, W\Vec{x_j}),
\end{equation}
\todo{shouldn't be multiplication by transpose of vector?}\todo[color=green]{No}
\begin{equation}
  \alpha_{ij} = \text{softmax}_j (e_{ij}),
\end{equation}
where $e_{ij}$ is the attention coefficients and $a$ represents a single-layer feed-forward neural network to perform self-attention on the nodes. The shared weight matrix $W$ is used for every node to perform linear transformation. $\alpha_{ij}$ indicates the importance of node \textit{j}'s features to node \textit{i} after normalizing $e_{ij}$ during the feature aggregation process.

\subsection{Graph pooling module}

To assign nodes to clusters at each hierarchical layer, we apply DIFFPOOL~\cite{ying2018hierarchical} to create  node embeddings and adjacency matrix for next coarsened layer (\textit{i+1}) from layer \textit{i}.
\begin{equation}
  Z^{(i)} = \text{GNN} (A^{(i)}, H^{(i)}),
\end{equation}
\todo{Replace $GNN$ with $\text{GNN}$}\todo[color=green]{done}
\begin{equation}
  (A^{(i+1)}, H^{(i+1)}) = \textsc{DIFFPOOL} (A^{(i)}, Z^{(i)}).
\end{equation}\todo{Replace with $\text{DIFFPOOL}$ or $\textsc{DiffPool}$}\todo[color=green]{done}

The graph pooling module takes the adjacency matrix $A^{(i)}$ and the features of the nodes or cluster nodes at layer $i$ as the input of the GNN module to get the new embedding matrices $Z^{(i)}$ of nodes or cluster nodes.  Then the DIFFPOOL module takes the node embedding matrices $Z^{(i)}$ and the adjacency matrix $A^{(i)}$ to generate a coarsened adjacency matrix $A^{(i+1)}$  and new embeddings $H^{(i+1)}$ for each of the nodes or cluster nodes in this coarsened graph. Then, the new coarsened graphs are fed to the GNN module to generate a coarser version of the input graph. This whole process is repeated several times until the final graph representation is generated, which contains only one general node or cluster node. Compared to other hierarchical representation learning methods, our model learns a hierarchical representation strategy automatically, which doesn't depend on the specific task and can be trained end-to-end. Generally, this unsupervised manner embeds the original graph to a coarser one by grouping the similar subgraphs together.

\subsection{Discriminator module}

Similar to Deep InfoMax~\cite{hjelm2018learning,velivckovic2018deep}, we introduce a discriminator module to help training the Encoder module and Graph pooling module, which enables our model to output the satisfied representations. The discriminator module trains the encoder to maximize the mutual information between a high-level graph representation and local features of the graphs and it is able to capture the unique graph representation for each graph individually. The local features are also included in the learned node embeddings, which represents the hierarchy of the original graphs. In this context, the final output representation of hierarchical learning is the graph-level summary representation $\Vec{s}$, and the local graph features are from the node embeddings of the  original graph $H = \{\Vec{h}_1, \Vec{h}_2, \ldots, \Vec{h}_n \}$. Therefore, our hierarchical model can be written by the following equation:
\begin{equation}
  \Vec{s} = \mathcal{R} (\text{GNN} (A, H)),
\end{equation}
\todo{Why does this function use $\text{GNN}(H,A)$ when the above GNN function is $\text{GNN}(A,H)$}\todo[color=green]{done}
where $A$ represents an adjacency matrix of the original graph, and $\mathcal{R} : \mathbb{R}^{n \times f} \rightarrow \mathbb{R}^{f'}$ is used to obtain a hierarchical graph-level representation. GNN module can be any node embedding module such as GCN and GAT. The readout function $\mathcal{R}$ utilizes the unsupervised hierarchical process to summarize  the graph-level vector $\Vec{s}$.

For the objective function, we follow the same loss function as DGI \cite{velivckovic2018deep}, which computes the standard binary cross-entropy between graph samples from the joint and the product of marginals:
\begin{align}
  \mathcal{L}  = \frac{1}{n + m} \Big(\sum_{i=1}^n \mathbb{E}_{(H, A)} [\log \mathcal{D} (\Vec{h}_i, \Vec{s})] + 
  \notag \\ 
  \sum_{j=1}^m \mathbb{E}_{(\Tilde{H}, \Tilde{A})} [\log (1- \mathcal{D}(\Vec{\Tilde{h}}_i, \Vec{s}))]\Big),
  \end{align}
  \todo{Replace $log$ with $\log$. What is $\mathbb{E}$?}\todo[color=green]{Expectation}
where a discriminator $\mathcal{D}: \mathbb{R}^f \times \mathbb{R}^f \rightarrow \mathbb{R} $, is employed to represent the probability scores of the local-global pair. The negative samples are drawn by combining the summary vector $\Vec{s}$ with the local features $\Vec{\Tilde{h}}_i$ from other graphs.  Through minimizing these log-expectation terms, our model can effectively extract useful local and global information of the input graph based on the mutual information maximization.

\section{Experiments}\label{section:experiments}

We evaluate the graph representation learned from \ProjectName~ on both graph classification and node classification tasks. In each case, \ProjectName~ is used to learn graph and node representations in a fully unsupervised manner. The graph and node classification tasks are performed by directly feeding the learned representations into simple linear classifiers. We also conduct the visualization experiments on learned representations to verify whether it's reasonable to assign clusters in an unsupervised manner.

\subsection{Data sets}

To evaluate the ability of \ProjectName~ to learn hierarchical representations from arbitrary complex graphs, we perform it on a variety of real-world graphs chosen from the commonly used benchmarks. For the node classification task, we consider the transductive learning setting and choose three standard data sets, Cora, Citeseer, and Pubmed~\cite{sen2008collective}, as summarized in Table~\ref{tab:node-dataset}. We employ the same training, validation and testing settings as those in DGI~\cite{velivckovic2018deep}, and report the node classification accuracy on the testing data, averaged over 50 runs of training. For graph classification task, we use protein data sets including D\&D~\cite{dobson2003distinguishing,shervashidze2011weisfeiler} and PROTEINS~\cite{borgwardt2005protein,dobson2003distinguishing}, the chemical molecules data set NCI1~\cite{wale2008comparison,shervashidze2011weisfeiler}, and the scientific collaboration data set COLLAB~\cite{yanardag2015deep}. More information on these data sets is shown in Table~\ref{tab:graph-dataset}. For this graph classification task, we perform 10-fold cross-validation to evaluate the performance, and apply the average over 10 folds as the final accuracy result. The visualization experiments are conducted on the data sets for graph classification tasks. We feed the original graph to output a coarser one based on the learned hierarchical cluster assignments.

\begin{table}
  \caption{Data set summary used in node classification task}
  \label{tab:node-dataset}
  \begin{tabular}{lllll}
    \toprule
    Data set&Nodes&Edges&Features&Classes\\
    \midrule
    Cora & 2,708 & 5,429 & 1,433 & 7\\
    Citeseer & 3,327 & 4,732 & 3,703 & 6\\
    Pubmed & 19,717 & 44,338 & 500 & 3\\
  \bottomrule
\end{tabular}
\end{table}

\begin{table}
  \caption{Data set summary used in graph classification task}
  \label{tab:graph-dataset}
  \begin{tabular}{lllll}
    \toprule
    Data set&Graphs&Classes&Avg.\# Nodes&Avg.\# Edges\\
    \midrule
    COLLAB & 5,000 & 3 & 74.49 & 2,457.78\\
    D\&D & 1,178 & 2 & 284.32 & 715.66\\
    PROTEINS & 1,113 & 2 & 39.06 & 72.82\\
    NCI1 & 4,110 & 2 & 29.87 & 32.30\\  
  \bottomrule
\end{tabular}
\end{table}

\subsection{Experimental setup}

As discussed in section~\ref{section:method}, \ProjectName~ includes encoder module, graph pooling module and discriminator module\todo{Don't capitalize Graph, Encoder, or Discriminator.}\todo[color=green]{done}. The encoder module encodes node representations using one GAT layer or one GCN layer. During the graph pooling module, we apply two DIFFPOOL layers to all of the data sets. Three GCN layers are performed between these two DIFFPOOL layers. In the hierarchical cluster setting, the number of clusters after DIFFPOOL layer is set be to 10-30\% of the number of nodes or clusters before pooling. The Readout function in the discriminator module is built on the top of the DIFFPOOL architecture, which enables us to learn the hierarchical graph representations.  Finally, the discriminator module relies on the mutual information maximization to achieve the unsupervised graph learning. We also apply Batch normalization~\cite{ioffe2015batch} after each layer. All models are trained for 1000 epochs with early stopping applied when the validation performance stops improving. We apply PyTorch framework~\cite{paszke2017automatic} to build graph neural network model and run it on NVIDIA Tesla V100 GPU. In order to demonstrate the effectiveness of our proposed model, we evaluate it on the following three tasks: node classification, graph classification,  and analysis of hierarchical cluster assignment.

\noindent {\bf Reproducibility: } Our source code and experimental data are available at \url{https://github.com/ifding/uhgr}.

\subsection{Results for Node Classification}

\begin{table*}
\centering
\caption{Node classification accuracies using different methods on datasets Cora, Citeseer and Pubmed. First column lists the type of data available during each graph representation learning method (X: node features, A: adjacency matrix, Y: node labels, X,A: unsupervised node representation learning, X,A,Y: supervised node classification).}\label{tab:node}
\begin{tabular}{llllll}
\toprule
Available data & Method & Cora & Citeseer & Pubmed\\
\midrule
X & Raw features & 47.9 $\pm$ 0.4\% & 49.3 $\pm$ 0.2\% & 69.1 $\pm$ 0.3\% \\    
A & DeepWalk & 67.2\% & 43.2\% & 65.3\% \\
X, A & DeepWalk + features& 70.7 $\pm$ 0.6\% & 51.4 $\pm$ 0.5\% & 74.3 $\pm$ 0.9\% \\
X, A & DGI & 82.3 $\pm$ 0.6\% & 71.8 $\pm$ 0.7\% & 76.8 $\pm$ 0.6\% \\   
X, A, Y & GCN & 81.5\% & 70.3\% & 79.0\% \\  
X, A, Y & GAT & 83.0 $\pm$ 0.7\% & 72.5 $\pm$ 0.7\% & 79.0 $\pm$ 0.3\% \\      
X, A & GAT-\ProjectName~ (ours) & 78.5 $\pm$ 0.1\% & 62.6 $\pm$ 0.3\% & 77.4 $\pm$ 0.6\% \\
X, A & GCN-\ProjectName~ (ours) & 76.7 $\pm$ 0.1\% & 62.5 $\pm$ 0.1\% & 75.1 $\pm$ 0.3\% \\
\bottomrule
\end{tabular}
\end{table*}

Table~\ref{tab:node} lists the node classification results on data sets Cora, Citeseer and Pubmed using our method and other existing methods. For the operation of node embeddings, we test two different GNN module variants: GATs and GCNs. The GATs module outperforms GCNs on most of the benchmarks, indicating that self-attention mechanism is more suitable for capturing local structural information. For the Cora and Citeseer data sets, we set both hidden dimension and output dimension to 320 and 400, respectively. And for the Pubmed data set,  128-dimensional hidden size and output size for GCN model and 100-dimensional hidden size and output size for GAT model are tested in our experiments. The node representations with larger hidden dimension and output dimension may be more powerful, and will be further optimized in future work.
\todo{You cannot use contractions in formal writing}\todo[color=green]{We could say that this is the problem deep learning studies which require powerful GPUs to train the models, I rewrite the above sentence.} 
According to the results, our model achieves better classification performance than DeepWalk, and obtains comparable performance with supervised learning methods.

\subsection{Results for Graph Classification}

\begin{table*}
\centering
\caption{Graph classification accuracies using different methods on datasets COLLAB, D\&D, PROTEINS and NCI-1. First column lists the type of data available during each graph representation learning method (X: node features, A: adjacency matrix, Y: node labels, X,A: unsupervised graph representation learning, X,A,Y: supervised graph classification).}\label{tab:graph}
\begin{tabular}{lllllll}
\toprule
Available data & Method & COLLAB & D\&D & PROTEINS & NCI-1\\
\midrule
X, A, Y & GRAPHSAGE & 68.3\% &75.4\% & 70.5\% & -\\    
X, A, Y & SET2SET & 71.8\% &78.1\% & 74.3\% & -\\
X, A, Y & DIFFPOOL & 75.5\% & 80.6\% & 76.3\% & 79.3\%\\   
X, A & graph2vec & - & - & 75.4 \% & 75.0 \% \\  
X, A & GAT-\ProjectName~ (ours) & 67.4\% & 75.6\% & 75.9\% & 66.6\%\\
X, A & GCN-\ProjectName~ (ours) & 66.9\% &77.4\% & 74.7\% & 66.6\%\\
\bottomrule
\end{tabular}
\end{table*}

Table~\ref{tab:graph} compares the graph classification performance of our unsupervised learning method with other supervised learning baselines on datasets COLLAB, D\&D, PROTEINS and NCI-1. The results show that our unsupervised method obtains similar performance as DIFFPOOL method on the PROTRINS benchmark and achieves comparable results with supervised methods, e.g. GRAPHSAGE, indicating that our method can learn useful graph representations even without graph labels. We also find that GAT-\ProjectName~ model performs better than GCN-\ProjectName~ model on the datasets COLLAB and PROTEINS, and performs worse than GCN-\ProjectName~ model only on the D\&D dataset. This suggests that different graph datasets need different Encoder layer to capture useful representations in order to achieve better classification performance. Compared with other unsupervised model, e.g., graph2vec~\cite{narayanan2017graph2vec}, GAT-\ProjectName~ model obtains comparable classification results on PROTEIN data set. However, graph2vec utilizes a SVM classifier to perform 1024-dimensional embeddings of graphs, where our method directly uses the graph representations to train and test a simple linear classifier. For the embedding dimensions, we simply set it to 20-360 to demonstrate the validity of the learned hierarchical representations, and doesn't further optimize this hyperparameter to achieve better classification performance due to hardware limitations.

\subsection{Visualization of hierarchical representation}

\begin{figure*}
\includegraphics[width=\linewidth]{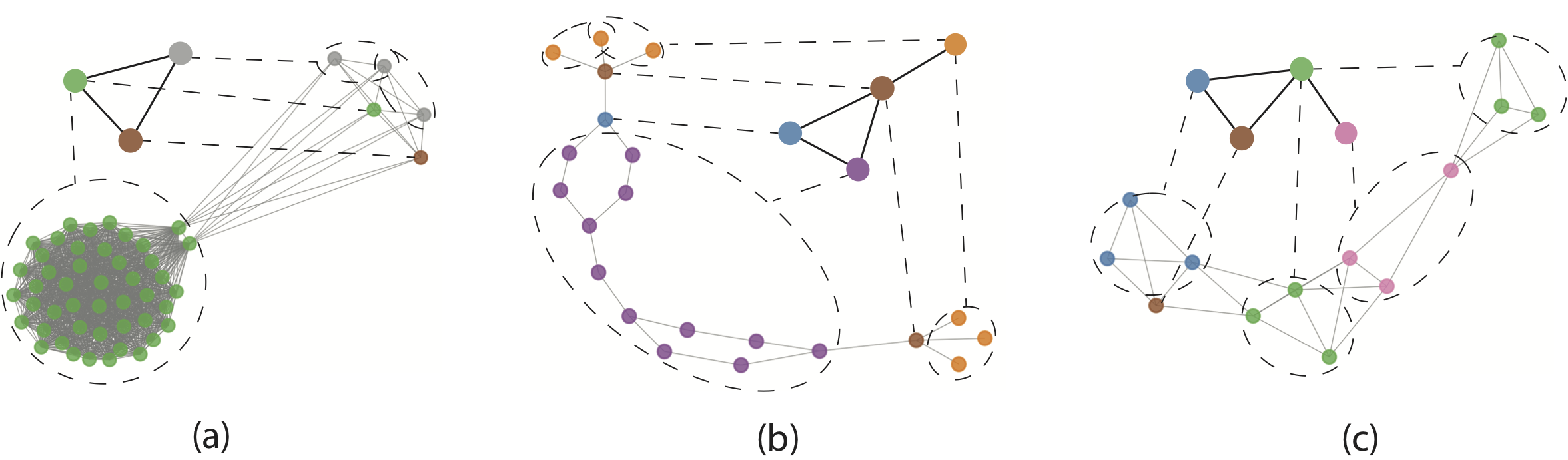}
\caption{ Visualization of hierarchical cluster assignment on data sets (a) COLLAB, (b) NCI1, (c) PROTEINS. The nodes of same color are merged into one cluster in the next layer and the dotted lines represents the cluster membership.}
\label{fig:visualization}
\end{figure*}

In addition to generating useful representation for classification tasks, our model can also create meaningful and interpretable representations in a hierarchical way. To evaluate the meanings of the learned hierarchical graphs, we visualize the cluster assignments after the DIFFPOOL layer. Figure~\ref{fig:visualization} shows the visualization of node assignments on the graphs from different data sets. Different node colors represent different node cluster labels from cluster assignment probabilities. Figure~\ref{fig:visualization} (a) is the node assignment on COLLAB data set and it is clear that our model can capture the hierarchical structure in these graphs. From  Figure~\ref{fig:visualization} (b) and (c), we also observe that many meaningful structures, including clique-like, tree-like and cycle-like structures, are captured by the model. This is because the DIFFPOOL layer computes the node assignment based on the node feature matrix and adjacency matrix, thus the input nodes with similar features and local structure obtain similar node assignment. Even if the subgraphs with similar patterns are far away, our model can still assign them into the same cluster. In general, our unsupervised learning method based on mutual information can capture different hierarchical structures.

\section{Related Work}\label{section:related}

\todo{Please delete last sentence. A reviewer will say this is incorrect.}\todo{IS: I'd delete the entire paragraph. Leaving only one sentence is not good.}\todo[inline, color=green]{done}

\textbf{Graph Neural Network.} 
A wide variety of graph neural networks have been applied in node classification~\cite{velivckovic2017graph,gao2018large,klicpera2018predict} and graph classification tasks~\cite{dai2016discriminative,duvenaud2015convolutional,zhang2018end,ying2018hierarchical,lee2018graph} in recent years. In node classification, GAT~\cite{velivckovic2017graph} stacks masked self-attentional layers to classify a node by attending \todo{what do you mean when you say ``by attending''? by traversing? and what is attending by different weights?}\todo[inline, color=green]{from the original paper, compute the representation of one node by its neighbors in different weights} over its neighbors in different weights. LGCN~\cite{gao2018large} builds a trainable graph convolutional layer to select a fixed number of neighboring nodes in order to transform graph data into grid-like data, which is suitable for typical convolutional operations. PPNP~\cite{klicpera2018predict} combines graph convolutional networks (GCN) and PageRank to overcome the problem that the size of the observed neighborhood of a node is difficult to extend.
In graph classification, the main challenge is to build a useful low-dimensional graph representation based on the node embeddings of the entire graph.
One straightforward solution, presented by Duvenaud \etal~\cite{duvenaud2015convolutional} and Veli{\u c}kovi{\' c} \etal~\cite{velivckovic2018deep} is to sum or average a graph's node embeddings. However, this solution ignores the structural information of graphs and considers that all nodes contribute the same weight to the calculation of graph representation.
Therefore, \textsc{DiffPool}~\cite{ying2018hierarchical} is proposed for graph classification that can learn hierarchical graph representations with a graph pooling module. Although this method solves the problem that existing GNN methods are flat and ignore hierarchical structure of graphs, it needs to learn under the supervision of graph-level labels. In addition, the real-world graphs are usually large and noisy, GAM~\cite{lee2018graph} is proposed for the attention-based graph classification, which utilizes the attention mechanism to focus on small but informative parts of graphs.
Combining local and global information on (hyper)graphs in the hierarchical setting has a long successful history. For example, in computational optimization domain, the multiscale solvers for (hyper)graph partitioning \cite{shaydulin2019relaxation}, separators \cite{hager2018multilevel}, and ordering \cite{ron2011relaxation} are among the top state of the art methods that preserve excellent time/quality trade-off.
However, all of these approaches depend on task-specific information to learn node embeddings or graph embeddings. In addition, most of them ignore the hierarchical representation of graphs, and thus have limited capabilities of capturing the natural structures of the real-world graphs~\cite{ying2018hierarchical}. \todo{You need to include a comparison between each method and how it is better/worse than the method you propose here.}\todo[inline, color=green]{the last two sentences summarize all of the above methods}

\textbf{Graph Representation Learning.}
Learning a high quality representation not only enables us to capture the latent variables of the data~\cite{bengio2013representation}, but also helps improve the performance of downstream tasks.
For graph-structured data, the learned low-dimensional representations (embeddings) can encode information of a graph's nodes, or the entire graph in the case of the GNN model. Many of the existing graph representations are focused on node embeddings by using random walk based objectives~\cite{perozzi2014deepwalk,grover2016node2vec,hamilton2017inductive}. In addition, LINE~\cite{tang2015line} and FOBE/HOBE~\cite{sybrandt2019fobe} focus on modelling first-order and second-order relationships between node neighborhoods to learn node embeddings and graph embeddings. VERSE~\cite{tsitsulin2018verse} is a simple graph embedding framework based on similarity measures. 
\todo{Need to cite VERSE if you're going to make this point.}\todo[inline, color=green]{done}
Glimer \etal~\cite{gilmer2017neural} propose a common framework to learn message passing algorithms and aggregate the node embeddings. Janossy Pooling~\cite{murphy2018janossy} is a permutation-invariant aggregator function to learn node embeddings.
Veli{\u c}kovi{\' c} \etal~\cite{velivckovic2018deep} propose an alternative unsupervised node embedding method based on mutual information~\cite{velivckovic2018deep}. HARP~\cite{chen2018harp} proposes a hierarchical paradigm to learn low-dimensional representations of a graph's nodes. This paradigm utilizes a smaller graph that approximates the original global structure to obtain good initializations for learning representations of the original graph.

Additional research focuses on learning representations of entire graphs in an unsupervised manner, which is quite different from the task of node embedding. In node embedding, the goal is to learn a low-dimensional vector to represent a node independently of supervised label information (\eg, node labels and graph labels). 
Graph2vec~\cite{narayanan2017graph2vec} is an unsupervised graph embedding method inspired by the document embedding models~\cite{le2014distributed}, but may not capture global structure, as this method only uses subtrees for graph embeddings.
Taheri \etal~\cite{taheri2018learning} generate sequences from graphs and train a long short-term memory (LSTM) autoencoder model to embed these graph sequences into continuous vectors. The LSTM network cannot be operated in parallel and is not appropriate to model large graphs. Some recent approaches have proposed applying the attention mechanism on graphs~\cite{choi2017gram,lee2018graph} that can determine which parts of the graph should have more attention. Yet the attention mechanism only focuses on local information which is not enough to achieve satisfactory node or graph representations. Recently, \textsc{BayesPool}~\cite{ubaru2020unsupervised} is proposed to use variational Bayes based on an encoder-decoder architecture to learn hierarchical graph representations in an unsupervised manner. Using Encoder-decoder architecture leads to this method being overly focused on node-based details, rather than more high-level node/graph embeddings. 
Different from previous representation learning methods, in this work we use an unsupervised learning framework based on mutuainformation with contrastive loss, to learn hierarchical graph representations.

\section{Conclusion}\label{section:conclusion}

In this paper, we propose an unsupervised hierarchical representation learning model based on mutual information, \ProjectName~, to learn node embeddings and graph embeddings. The mutual information maximization between global representation and local parts of the graphs can encourage the model to learn related structural information in all locations. This unsupervised learning model is able to learn task-independent graph representations. In addition, it can learn hierarchical graph representation, which is meaningful and easy to interpret. To demonstrate the effectiveness of the model, we perform node classification and graph classification tasks based on the learned representations. The results show that our \emph{unsupervised} model can achieve comparable results with the \emph{supervised} methods on several tested data sets. Finally, through visualization of the hierarchical cluster assignment, we show that our model is able to generate hierarchical representations by clustering different meaningful structures which increases interpretability.

%

%
\bibliographystyle{ACM-Reference-Format}
\bibliography{sample-base}

%









\end{document}